\documentclass[conference]{IEEEtran}
\usepackage{cite}

\ifCLASSINFOpdf
  \usepackage[pdftex]{graphicx}
  \graphicspath{{../pdf/}{../jpeg/}}
  \DeclareGraphicsExtensions{.pdf,.jpeg,.png}
\else
  \usepackage[dvips]{graphicx}
  \graphicspath{{../eps/}}
  \DeclareGraphicsExtensions{.eps}
\fi

\usepackage[cmex10]{amsmath}

\usepackage{array}
\usepackage{url}

\usepackage{multirow}
\usepackage[tight,footnotesize]{subfigure}

\begin{document}

\title{Reduced Memory Region Based Deep\\Convolutional Neural Network Detection}

\author{
\IEEEauthorblockN{Denis Tom\'e, Luca Bondi, Luca Baroffio, Stefano Tubaro}
\IEEEauthorblockA{Dipartimento di Elettronica, Informazione e Bioingegneria\\
Politecnico di Milano\\
Milano, Italy\\
denis.tome@mail.polimi.it, luca.bondi@polimi.it,\\luca.baroffio@polimi.it,
stefano.tubaro@polimi.it}
\and\IEEEauthorblockN{Emanuele Plebani, Danilo Pau}
\IEEEauthorblockA{Advanced System Technology\\
STMicroelectronics\\
Agrate Brianza, Italy\\
emanuele.plebani1@st.com, danilo.pau@st.com}
}
\maketitle

\begin{abstract}
Accurate pedestrian detection has a primary role in automotive safety: for 
example, by issuing warnings to the driver or acting actively on car’s brakes, 
it helps decreasing the probability of injuries and human fatalities. In order 
to achieve very high accuracy, recent pedestrian detectors have been based on 
Convolutional Neural Networks (CNN). Unfortunately, such approaches require 
vast amounts of computational power and memory, preventing efficient 
implementations on embedded systems. This work proposes a CNN-based detector, 
adapting a general-purpose convolutional network to the task at hand. By 
thoroughly analyzing and optimizing each step of the detection pipeline, we 
develop an architecture that outperforms methods based on traditional image 
features and achieves an accuracy close to the state-of-the-art while having 
low computational complexity. Furthermore, the model is compressed in order to 
fit the tight constrains of low power devices with a limited amount of 
embedded memory available. This paper makes two main contributions: (1) it 
proves that a region based deep neural network can be finely tuned to achieve 
adequate accuracy for pedestrian detection (2) it achieves a very low memory 
usage without reducing detection accuracy on the Caltech Pedestrian dataset. 
\end{abstract}

\begin{IEEEkeywords}
Window proposals, CNN, Object Detection, fine tuning, embedded systems
\end{IEEEkeywords}

%
\IEEEpeerreviewmaketitle

\section{Introduction}
Figures form NHTSA’s Fatal Analysis Reporting System (FARS) in 2014 show 
that 32,675 
people died in motor vehicle crashes and the fatality rate for 2015 is 
estimated to reach 1.17 deaths per 100 million vehicle miles traveled 
\cite{NHTSA2016}. These numbers show the importance of building automated 
vision systems for pedestrian and car detection. As the world urbanizes more 
and more, accidents involve 33K lives, 250K disabilities and 2M injuries
accounting for \$300B of damage. In 95\% of the 
cases human error is the cause, mostly by passenger distraction or changes in
traffic / road / environmental conditions realized too late. This situation
calls
for urgent actions by the automotive industry in order to react and propose
advanced
safety measures to the driver, and visual object detection is instrumental to
that need.

The most successful and accurate approaches in object detection are based on 
convolutional neural networks (CNN) which have significantly outperformed
methods based on densely extracted features \cite{Hosang2015a}. CNNs
integrate the feature extraction and feature classification stages of an object 
detector in an end-to-end approach by training the model parameters on a large
dataset. However, the resulting models are characterized by a large 
computational complexity and number of parameters: for example, the successful
AlexNet model \cite{Krizhevsky2012} requires one billion floating point
operations per classification and 60 million parameters, corresponding to
217\,MB of parameters memory in single precision floating point. The more
accurate VGG networks \cite{Simonyan2014a} can require up to 40 billion floating
point operations and 500\,MB of parameter memory.

The complexity of CNNs prevents performing classification at every potential
position and scale and thus a widely used approach, first proposed in 
\cite{Girshick2013}, is to use a object proposal mechanism, where only
``object-like'' regions are processed by the network. The computation can be
further sped-up by using Graphic Processing Units (GPU) or specialized hardware
(such as \cite{Han2016}). However, the memory required for the parameters puts
severe
constraints on embedded platforms, where a larger amount of on-chip memory
increases costs and access to an external Dynamic Random Access Memory (DRAM)
requires two orders of magnitude more energy than accessing local Static RAM
(SRAM) caches \cite{Han2015a}. This motivates developing a fully embedded
memory implementation of CNN, where suitable compression schemes are applied
to the network weights while minimizing the loss in accuracy.

This paper proposes a set of strategies tailored at significantly reducing
the amount of space required by the parameters of a convolutional neural
network. Starting from \emph{DeepPed}, an optimized pedestrian detection
pipeline we
previously developed \cite{Tome2015}, we reduce the redundancy of the parameters
with two approaches inspired by \cite{Han2015a}: by compressing the individual
weights through \emph{k}-means quantization, and by pruning weights with small
absolute value. We evaluate both approaches separately and in combination, in
order to find the best trade-off between accuracy and memory requirements.

The rest of this paper is organized as follows. In section \ref{sec:sota} a 
review of the state of the art about neural network compression is offered to 
the reader. Sections \ref{sec:pipe} and \ref{sec:compr} illustrate respectively 
the proposed detection pipeline and the model compression scheme. In section 
\ref{sec:results} numerical results from experiments on the Caltech Pedestrian
dataset are presented and discussed. Finally, in section \ref{sec:conclusions}
some conclusions are drawn.

\section{Review of the state of the art}
\label{sec:sota}

\begin{figure*}[!t]
\centering
  \subfigure[]{
    \includegraphics[width=0.40\textwidth]{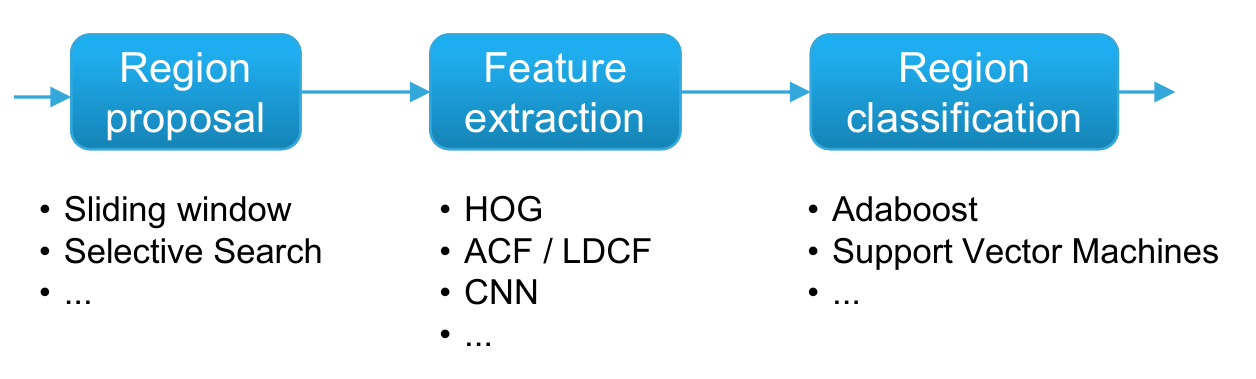}
    \label{fig:pipe}
  }
  \subfigure[]{
    \raisebox{4mm}{\includegraphics[width=0.52\textwidth]{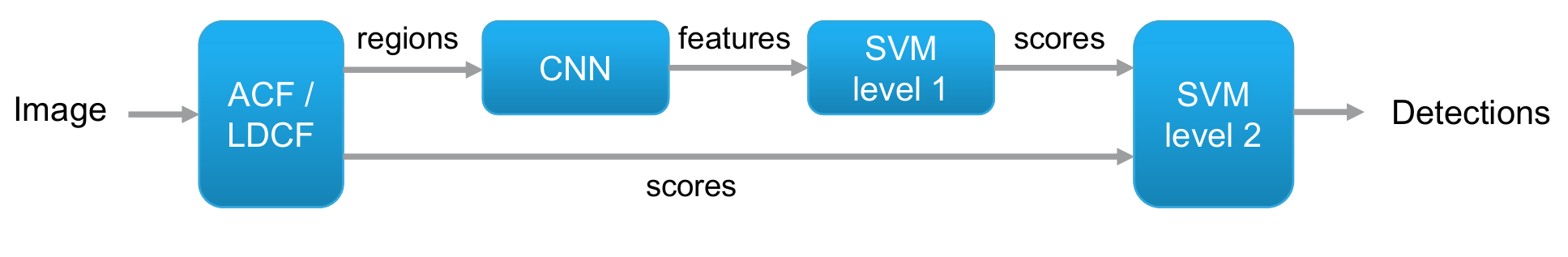}}
    \label{fig:pipe_cnn}
  }
  \caption{Detector pipelines. In (a), a typical pedestrian detection pipeline 
in the state of the art methods. In (b), proposed pipeline for Region based 
Convolutional Neural Networks (R-CNN).}
\end{figure*}

Starting with the introduction of AlexNet in 2012, which won by a large margin 
the ImageNet Large Scale Visual Recognition Challenge \cite{Krizhevsky2012},
deep convolutional neural networks have become the dominant approach in image 
classification, recently achieving human-level performances \cite{He2015}.
The seminal work by Girschick on Regions with convolutional neural networks 
(R-CNN) \cite{Girshick2013} extended those results to object 
detection. In the R-CNN paradigm, detection is performed in two stages: 
first, a separate algorithm generates candidate object proposals based on e.g. 
region segmentation or edges; then, a CNN is run on the proposals to generate 
the final detection results. While for generic object detection an object
agnostic method such as Selective Search \cite{Uijlings2013} is necessary, for
more focused tasks like pedestrian detection such proposals are
under-performing compared to the state of the art 
\cite{Hosang2015a} and by using an accurate pedestrian detector as proposal 
method, significant gains can be achieved. Moreover, the proposals generated by 
a specialized object detector have a higher probability of containing the 
object of interest, reducing the number of proposal regions needed: as shown 
by \cite{Hosang2015a} in the supplemental material, only 3 proposal per images 
are enough to reach a recall of more than 90\%.
While methods for neural network compression have a long history 
\cite{LeCun1990}, the pace of research has accelerated in response to the large 
networks introduced after 2012. Denil \emph{et al.} demonstrated that the
parameters of deep neural networks are highly redundant
and can be reduced up to 20$\times$ with no appreciable loss of accuracy, giving
a strong incentive to network compression.

Early approaches are based on enforcing weight 
sparsity, either through low-rank approximations \cite{Denton2014} or by an 
opportune regularization term \cite{Collins2014} \cite{Liu2015a}, achieving a 
compression ratio up to 20$\times$ at the price of a 1-2\% loss of accuracy.
The convolutional layers of the network can be compressed either by 1-rank
approximation \cite{Jaderberg2014}, filter decomposition \cite{Zhang2015b} or
Tucker decomposition \cite{Kim2015}, achieving a compression rate of
5$\times$. The sparsity of convolutional layers has the added benefit of
reducing the number of operations, up to 4$\times$ in the case of
\cite{Liu2015a}.
Other approaches replace the fully connected layers of the network (the ones 
contributing the most to the number of parameters) with a different kind of 
layer with a lower number of parameters. Examples are
kernel machines \cite{Yang2014a}, tensors \cite{Novikov2015}, circulant 
matrices \cite{Cheng2015} or a cascade of diagonalized matrices
\cite{Moczulski2015}; convolutional layers can also be replaced by 
separable filters \cite{Jin2014a}. Alternatively, the network weights can be
compressed by an hashing function and the hashing trick used to carry out
computations directly in the compressed space.
Stages with $1\times1$ convolutions are an effective way of reducing the number
of parameters in a network and they are 
often used in very deep networks such as the Inception architecture 
\cite{Szegedy2014} and in Residual Networks \cite{He2015a}; however,
when applied to smaller networks can achieve the same level of accuracy as
AlexNet with 40$\times$ less parameters \cite{Iandola2016}.


However, a simple strategy based on scalar quantization of the weights 
\cite{Gong2014} and connection pruning \cite{Han2015} is surprisingly effective 
and with network retraining achieves a 37$\times$ compression on AlexNet with
almost no loss in performance \cite{Han2015a}. The performances of this
approach are further improved by enforcing a layer-wise reconstruction penalty
to the quantized weights \cite{Wu2016}.

\section{Proposed pipeline}
\label{sec:pipe}

\enlargethispage{-\baselineskip}

The baseline CNN detector, dubbed \emph{DeepPed} \cite{Tome2015}, follows from
\cite{Hosang2015a} in combining R-CNN with an efficient pedestrian detector
used as proposal method. The Aggregated Channel Features (ACF) detector
\cite{Dollar2014a} has been chosen for its speed and accuracy. An improved
version of ACF, known as
Locally Decorrelated Channel Features (LDCF) \cite{Nam2014}, was also taken into
account, but despite its higher accuracy it was discarded due to its 
computational complexity, which is ten times higher than ACF.

In \emph{DeepPed}, an 
input image is analyzed by ACF and several regions are proposed as potential 
pedestrians, with a score associated to each region. A pre-trained AlexNet
network, trained on the 1000 categories of the ImageNet challenge and publicly
available\footnote{BVLC AlexNet Model in Caffe:
\url{https://github.com/BVLC/caffe/tree/master/models/bvlc_alexnet}},
is used as starting point. The last classification layer, which is
application-specific, is removed and and replaced with a Support Vector Machine
(SVM). The model is then retrained by fine-tuning on the
Caltech Pedestrian training dataset \cite{Dollar2012} sub-sampled by 3$\times$
as suggested by \cite{Dollar2014a} and using 6-fold cross-validation
to select the best performing network. The training examples are pedestrian and
non-pedestrian windows chosen by ACF and selected among the highest-scoring
proposals. The ACF score and the SVM score from the CNN are then further
combined by stacking a second SVM trained on the validation set, which gives
the final detection score.

\begin{figure}[t]
\centering
\includegraphics[width=3.5in]{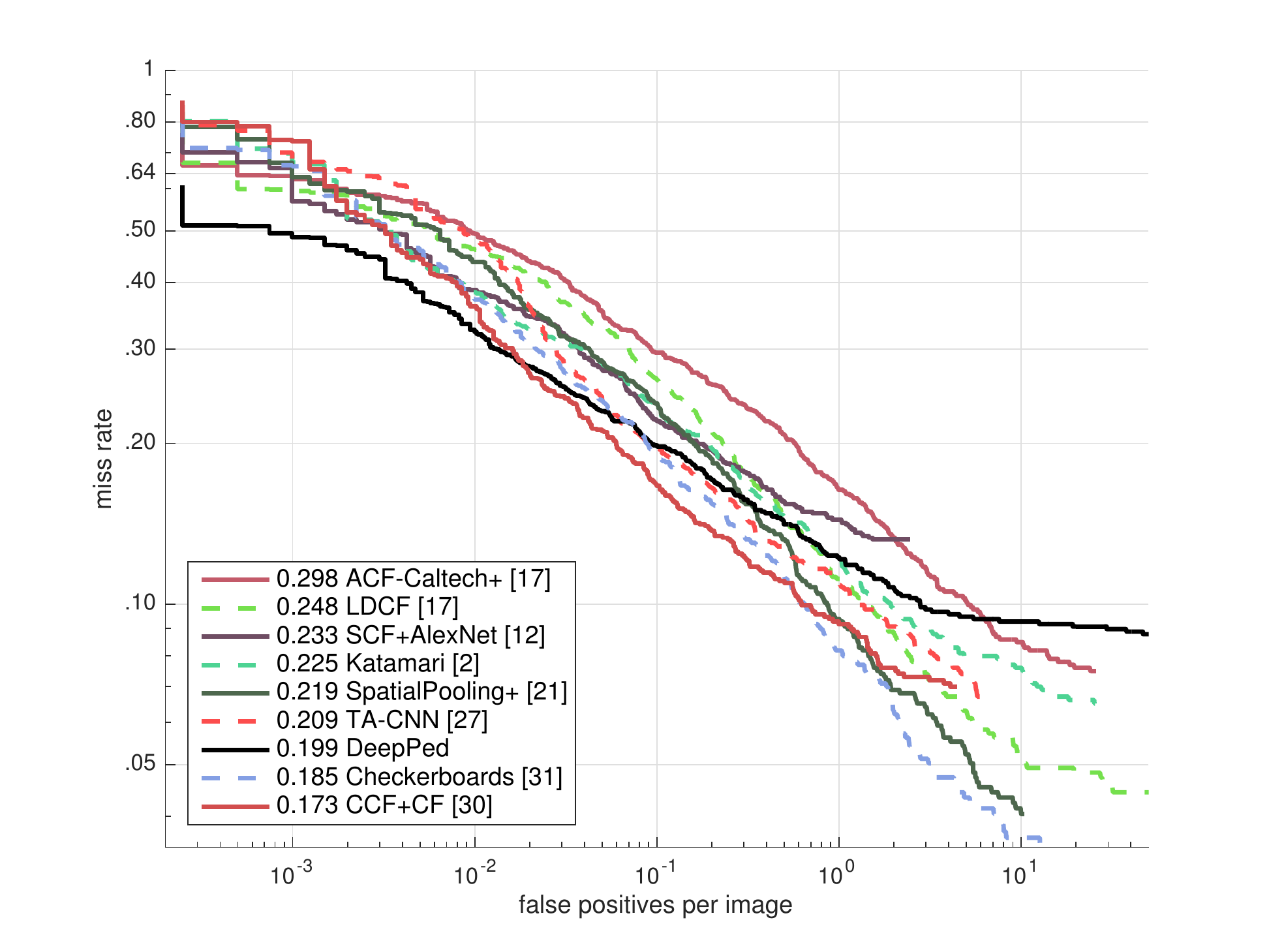}
\caption{Comparison between the proposed \emph{DeepPed} (solid black line) and other 
popular pedestrian detection algorithms. Note that the proposed method does not
make use of multiple frame information, i.e. it does not exploit the optical
flow between contiguous frames.}
\label{fig:deep_ped}
\end{figure}

Figure \ref{fig:deep_ped} shows the performance of the final \emph{DeepPed} pipeline
compared with other state-of-the art approaches in pedestrian detection.
The details of the algorithm are discussed in \cite{Tome2015}, 
where each step of the pipeline is analyzed and optimized in order to increase 
the final detection accuracy. In the following sections, we focus on reducing
the space required to store the parameters of this network.

\section{Model compression}
\label{sec:compr}

In order to reduce the parameter memory, two strategies inspired by 
\cite{Han2015a} have been chosen to compress the network weights: 
\emph{scalar quantization} and \emph{weight pruning}. In the
first case, we consider for each layer the distribution of individual weights
and we quantize their values with \emph{k}-means using a variable number of
centroids; in the second case, we set to zero the weights with the lowest
absolute value, using different values for the threshold in order to change the
proportion of non-zero weights. Finally, the two approaches are combined,
either by quantizing and then pruning the resulting weights, or by pruning and
then quantizing the resulting distribution. As the experiments in Section
\ref{sec:results} will show, the two approaches are largely independent one 
from the other and thus the compression factors can be composed maintaining
roughly the same accuracy level.

More in detail, the following procedures have been used to compress the weights:

\begin{itemize}
\item \textbf{Scalar quantization}: each CNN layer is compressed individually.
All the weight values in the layer parameters are clustered using 
the \emph{k}-means algorithm, where the number of centroids is chosen as a
function of the compression factor. Assuming that the uncompressed weights are 
represented each with $B$ bits, the number of centroids is: 

\begin{equation*}
n_{centroids} = 2^{\frac{B}{f_{compr}}}
\end{equation*}

Henceforth, we will consider a single precision floating point representation
($B = 32$) for the uncompressed weights. The maximum achievable compression
rate is thus $32$ ($n_{centroids} = 2$).

\item \textbf{Pruning}: each CNN layer is compressed individually. Weights
whose absolute value is smaller than a threshold are zeroed out, i.e.:

\begin{equation*}
w_{i,j} =\begin{cases}
      0, & \text{if}\ \left| w_{i,j} \right| < threshold \\
      w_{i,j}, & \text{otherwise}
    \end{cases}\end{equation*}

The threshold is set as the $p$th percentile of the weight distribution, where
$p = 100 \cdot (1 - \frac{1}{f})$ and $f$ is the compression factor; i.e. the
75th percentile is chosen for $f = 4$.

\item \textbf{Quantization and Pruning}: The combination of both quantization 
and pruning lets the strengths of one method compensate the shortcomings of the 
other.
As shown in \ref{sec:results}, a higher level of compression can be achieved 
with respect to the single methods alone, while reaching the same level of accuracy.
\end{itemize}

Differently for quantization, the distribution of pruned connections needs to
be stored, increasing the storage 
requirements. For example, if a \emph{binary map} coding the pruned or
not-pruned status of the connections is used, the fully connected layers of
\emph{DeepPed} would require a map of 6.8\,MB. However, in a scenario in which a
hardware
designer can draw on an Application Specific Integrated Circuit (ASIC) or on an
Field Programmable Gate Array (FPGA) all the connections between neurons
individually, the pruned weights would translate in missing connections,
and since power consumption increases linearly 
with wire capacity loads, the pruned connections would reduce power, space and 
storage usage.

\begin{figure*}[!tb]
\centering
  \subfigure[]{
    \raisebox{2mm}{\includegraphics[width=0.45\textwidth]{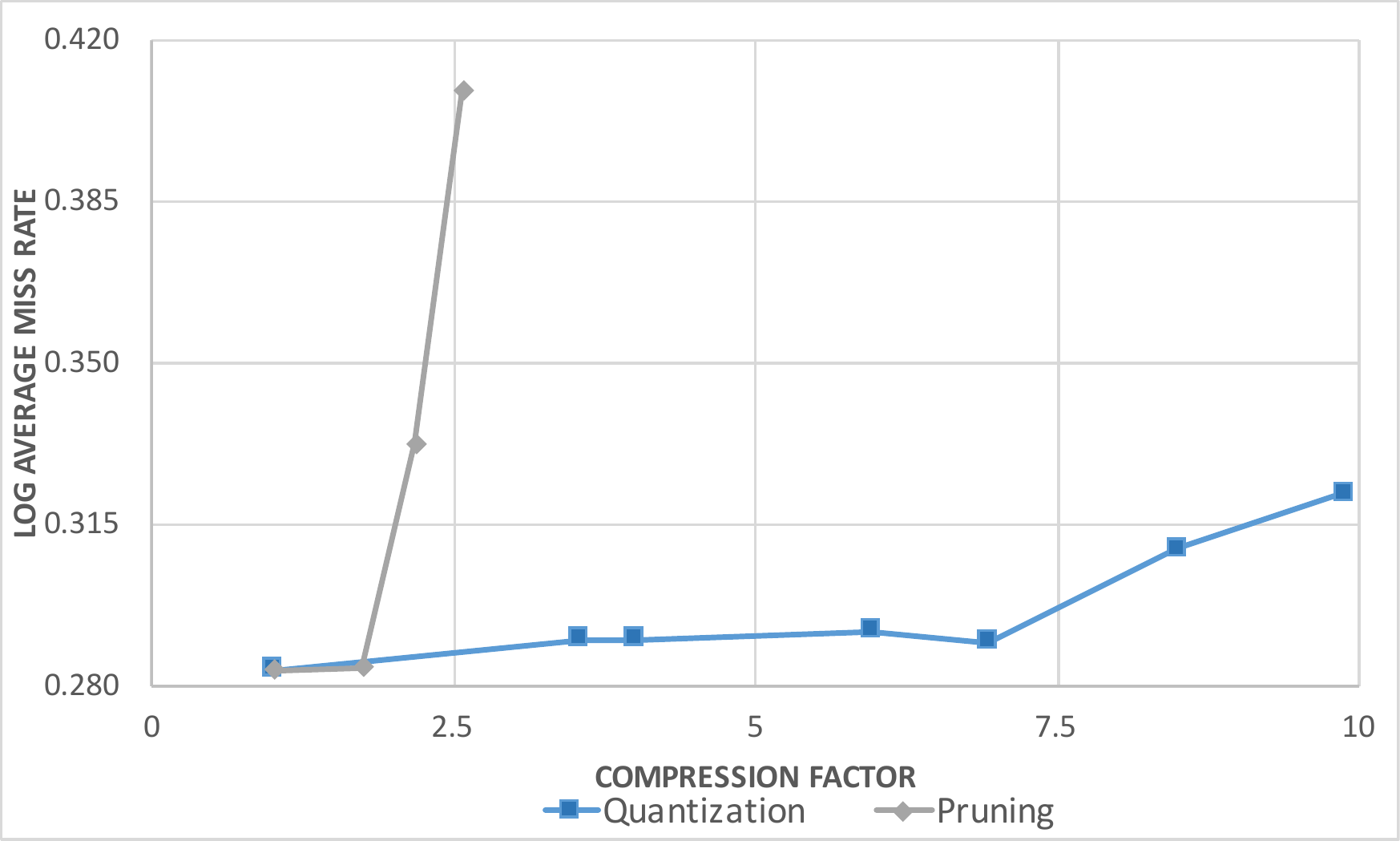}}
    \label{fig:compr_conv}
  }
  \subfigure[]{
    \includegraphics[width=0.45\textwidth]{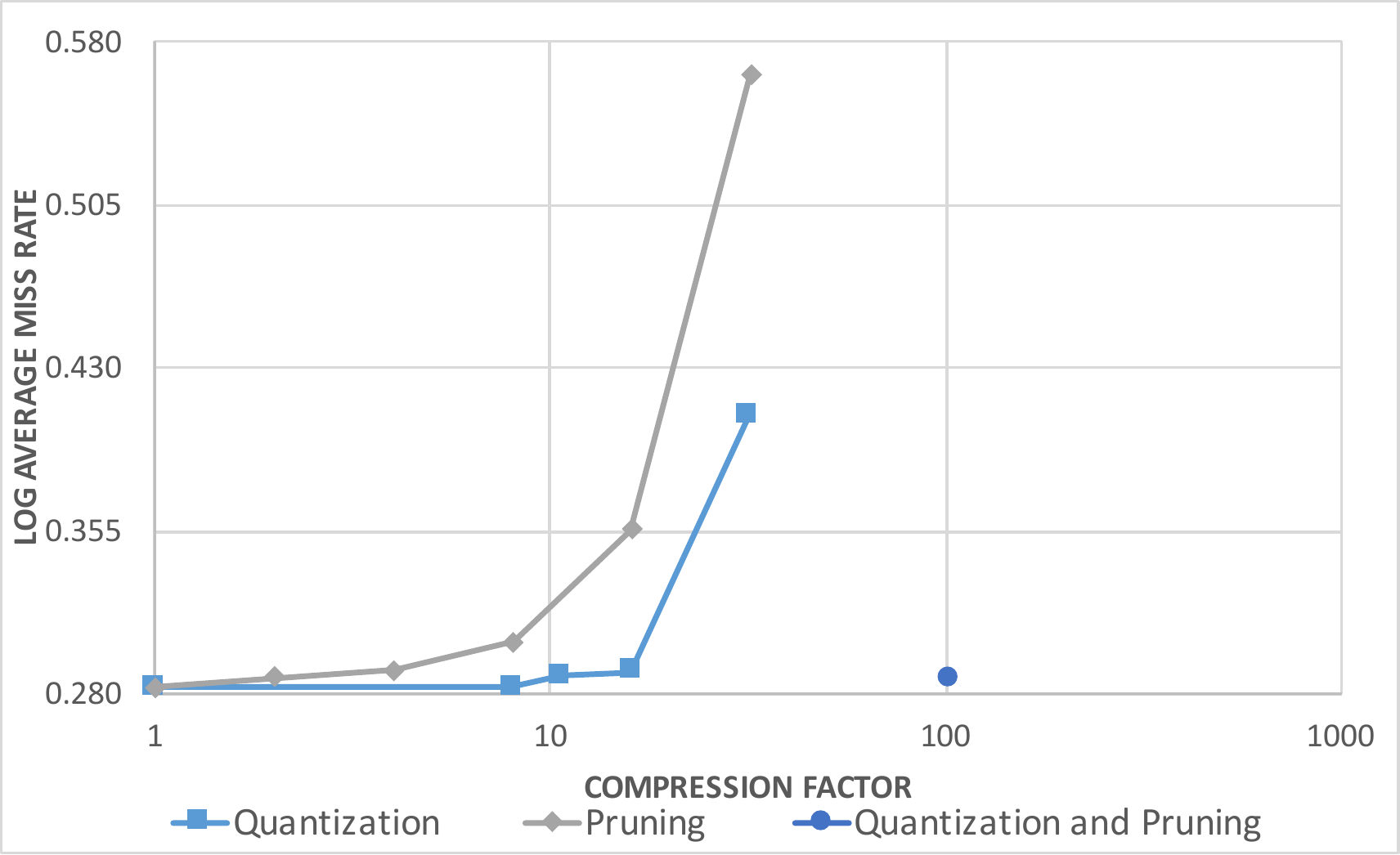}
    \label{fig:compr_fc}
  }
  \caption{\emph{DeepPed} results. In (a), comparison between compression by 
scalar quantization and by pruning in convolutional layers: the pruning 
method has a dramatic impact on the information content. In (b), comparison
between compression by scalar quantization and by pruning in fully connected 
layers.}
\end{figure*}

\section{Experimental evaluation and results}
\label{sec:results}

\begin{table*}[!bt]\footnotesize
\centering
\renewcommand{\arraystretch}{1.1}
\caption{Final compressed model. Compression factors and sizes per layer.}
\label{tab:compr}
  \begin{tabular}{ l | c  c  c  c }
    \hline
    \textbf{Layer} & \textbf{CF Quantization} & \textbf{CF Pruning} & 
         \textbf{Original size (MB)} & \textbf{Compressed size (MB)}\\ \hline
    \emph{Conv 1} & 3.56  & - & 0.13 & 0.037 \\
    \emph{Conv 2} & 4 & - & 1.17 & 0.293 \\
    \emph{Conv 3} & 4 & 2 & 3.38 & 0.422 \\
    \emph{Conv 4} & 4 & 2 & 2.53  & 0.319 \\
    \emph{Conv 5} & 4 & - & 1.69  & 0.422 \\ \hline
    \emph{Fully Connected 1}  & 16 & 4 & 144.02 & 0.66 \\
    \emph{Fully Connected 2}  & 16 & 4 & 64.02 & 1.35 \\  \hline
    \emph{Total} & \multicolumn{2}{c}{61.92} & 216.94 & 3.50 \\
  \end{tabular}
\end{table*}

In the experiments, we tested the following approaches:

\begin{itemize}
\item scalar quantization alone
\item weight pruning alone
\item weight pruning followed by scalar quantization
\end{itemize}

Moreover, we tested the compression separately on two different set of layers:

\begin{itemize}
\item compressing the convolutional layers
\item compressing the fully connected layers
\end{itemize}

Most of the network parameters reside in the fully connected layers, and thus
the latter set is the most important; however, to reach high compression 
rates, both sets of layers need to be taken into account. The target size
is 4\,MB, a typical size for local SRAM in embedded platforms.

To assess the impact of compression on the network accuracy we resort to the
evaluation metrics proposed by Doll\'ar \emph{at al.} 
\cite{Dollar2012} on the Caltech Pedestrian dataset. In particular, the 
performance of the different methods is evaluated in terms of trade-off 
between the \emph{compression factor} (CF) and the \emph{log average miss rate} 
(LAMR), as measured on the Caltech Pedestrian test set. The LAMR metric computes
the geometric mean of the miss rate in the interval between 0.01 false
detections per frame and 1 false detections per frame and can be interpreted as
a smoothed
estimate of the miss rate at 0.1 false detections per frame. Besides the initial
fine-tuning of the \emph{DeepPed} uncompressed model, the methods we tested 
do not require additional training, so only the test set has been used 
in the experiments. The original \emph{DeepPed} model with ACF proposals reaches a
LAMR of 28.3\%.

We assess the effect of scalar quantization by measuring the accuracy of the
model after compressing the fully connected (\emph{fc}) layers at a factor of
8, 10.7, 16 and 32 (4, 3, 2 and 1 bit per weight) and keeping the convolutional
layers unchanged. Likewise, we assess pruning by reducing the number of
connections by 2, 4, 8, 16 and 32 times. The results are shown in Figure
\ref{fig:compr_fc}, with the compression factors in logarithmic scale: 
quantization is more efficient than pruning, and compression
up to 8$\times$ can be achieved without appreciable loss of accuracy.

In the case of convolutional (\emph{conv}) layers, we observe that the first
layer is strongly affected by compression, while the fourth and fifth layers
are more resilient. For this reason, the first layer is always compressed with
9 bits (CF=3.56) and never pruned; the remaining layers are compressed or 
pruned with increasingly high factors. Scalar quantization is tested at a factor
of 3.56, 4, 6, 7, 8.5 and 10; pruning is tested by using the factors 
$\left(1, 2, 2, 2 \right)$, $\left(2, 2, 2, 4 \right)$ and
$\left(2, 2, 4, 4 \right)$ for the convolutional layers from 2 to 5, resulting 
in overall compression factors for convolutional layers at 1.74, 2.17 and 2.57.
The results are shown in Figure \ref{fig:compr_conv}: the performances degrade
much faster than in the case of
fully connected layers, as expected from the fact that weight sharing in the
convolutional layers already counts as a form of parameter reduction. As in the
case of fully connected layers,
convolutional layers are more robust to scalar quantization, achieving
compression factor up to 6$\times$ with a small cost in accuracy. Pruning 
instead leads to a rapid degradation of performances with factors greater than
2$\times$, showing that throwing away weights with small coefficients is not
desirable in convolutional layers, since the effect of these small
contributions greatly influences the final accuracy.

The two methods are combined in the case of fully connected layers, where
we prune $\frac{2}{3}$ of the connections and we quantize the remaining weights
with 1 bit, resulting in a compression factor of 102$\times$. As Figure
\ref{fig:compr_fc} shows, combining the two approaches actually helps both,
and despite a difference of an order of magnitude in compression factor, the 
accuracy is comparable to scalar quantization at 10$\times$. Pruning even
improves the results of quantization at 1 bit, because now the centroids need
not to fit irrelevant weights. By compressing only the fully connected weights,
the model already reaches a size of 10.9\,MB.

\vspace{-2mm}
\begin{table}[b]\footnotesize
\centering
\renewcommand{\arraystretch}{1.1}
\caption{Best performing compressed models.}
\label{tab:sumcompr}
  \begin{tabular}{  l | c  c | c  c | c }
    \hline
    \multirow{2}{*}{\textbf{Layers}} & \multicolumn{2}{c|}{\textbf{Compression}} & 
    \multicolumn{2}{c|}{\textbf{Size (MB)}} & \multirow{2}{*}{\textbf{LAMR}} \\
     & \textbf{Quant.} & \textbf{Pruning} & 
       \textbf{Original} & \textbf{Compr.}\\ \hline
    \emph{conv1-5} & \multicolumn{2}{c|}{5.94$\times$} & 2.23 & 1.5 & 29.2\% (+0.9\%) \\
    \emph{fc6-7}  & \multicolumn{2}{c|}{102$\times$} & 208.04 & 2.04 & 28.6\% (+0.3\%) \\ \hline
    \emph{Total} & \multicolumn{2}{c|}{61.35$\times$} & 216.93 & 3.54 & 28.7\% (+0.4\%) \\
  \end{tabular}
\end{table}

We finally combine compression of convolutional and fully connected layers.
Table \ref{tab:sumcompr} summarizes
the performances for the best model in each scenario and for the final selected
model; for the final model, Table \ref{tab:compr} shows the compression factors
and sizes layer by layer. The result is a model with a total size of 3.5\,MB and 
an overall compression factor of 61.92.
The final model accuracy is only slightly worse than the accuracy reachable when compressing only the fully connected layers (LAMR from $28.6\%$ to $28.7\%$), despite requiring less than one third the size of the latter.
Moreover, when fully connected compression is applied together with 
convolutional compression, the overall accuracy increases with respect to the case when only convolutional layers are compressed (LAMR from $29.2\%$ to $28.7\%$). A possible motivation for this behavior is related to the ``screening'' capabilities of quantization applied on fully connected layers, acting as a  filter on the noise generated in convolutional layers compression.
Since 4\,MB of embedded memory are viable in the 28\,nm Fully Depleted Silicon
On Insulator (FDSOI) STMicroelectronics fabrication 
process, this model enables several low-power and low-cost embedded
applications.

The \emph{DeepPed} architecture combined with the proposed compression scheme
has been ported on an NVIDIA
Jetson TK1 board and integrated with an optimized implementation of the ACF
detector. The detector is capable of running at 2.4 frames per second (fps)
when processing 5 proposal per frame. Moreover, by applying
a tracking-by-detection algorithm such as \cite{Breitenstein2011} and using an
average tracking length of 5 frames, the speed of the detector increases up to
10 fps, most of it spent in the CNN evaluation stage.

\section{Conclusions}
\label{sec:conclusions}

In this paper we present a detailed study of the effect of neural network
compression in the case of a pedestrian detector and we show that a
combination of simple yet effective methods allows to significanlty reduce the
memory required to store the network parameters, achieving a final compression
factor of 62$\times$. The behavior analysis of both convolutional and fully 
connected layers under scalar quantization and connection pruning shows 
that while the fully connected layers are the most redundant ones, high
compression rates are achievable on convolutional layers with a small effect on
the final accuracy. Thus, with a proper choice of compression parameters the
accuracy of the system is preserved while allowing the development of embedded 
architectures at a small cost and low power consumption.

\enlargethispage{-2\baselineskip}

\bibliographystyle{IEEEtran}
\bibliography{IEEEabrv,paper}
%



\end{document}